\documentclass[11pt]{article}
\usepackage{paclic34}
\usepackage{times}
\usepackage{latexsym}
\usepackage{comment}
\usepackage{amsmath}
\usepackage{graphicx}
\usepackage{multirow}
\usepackage{url}
\usepackage{color}

\title{A Simple Disaster-Related Knowledge Base for Intelligent Agents}
\author{Clark Emmanuel Paulo, Arvin Ken Ramirez, David Clarence Reducindo \\ \textbf{Rannie Mark Mateo, Joseph Marvin Imperial}\\
  National University \\
  Manila, Philippines \\
  {\tt {jrimperial}@national-u.edu.ph} \\}

\begin{document}

\maketitle

%%
%% --- Abstract --- %%
%%

\begin{abstract}
In this paper, we describe our efforts in establishing a simple knowledge base by building a semantic network composed of concepts and word relationships in the context of disasters in the Philippines. Our primary source of data is a collection of news articles scraped from various Philippine news websites. Using word embeddings, we extract semantically similar and co-occurring words from an initial seed words list. We arrive at an expanded ontology with a total of 450 word assertions. We let experts from the fields of linguistics, disasters, and weather science evaluate our knowledge base and arrived at an agreeability rate of 64\%. We then perform a time-based analysis of the assertions to identify important semantic changes captured by the knowledge base such as the (a) trend of roles played by human entities, (b) memberships of human entities, and (c) common association of disaster-related words.  The context-specific knowledge base developed from this study can be adapted by intelligent agents such as chat bots integrated in platforms such as Facebook Messenger for answering disaster-related queries.
\end{abstract}

%%
%%
%% --- 1 Introduction --- %%
%%
%%

\section{Introduction}

% * * * 
% Papers for PACLIC should focus on Language (intricacies) + Computation (challenges in processing and representing language)
% * * *

The Philippines is a common ground for natural disasters such as typhoons and flooding. According to the latest statistics of Philippine Atmospheric, Geophysical and Astronomical Services Administration (PAG-ASA)\footnote{Statistics on annual tropical cyclones in the Philippines: http://bagong.pagasa.dost.gov.ph/climate/tropical-cyclone-information}, there are more tropical cyclones entering the vicinity of the Philippines than anywhere else in the world. In a year, almost 20 tropical cyclones enter with 70\% chance of developing into a full-blown typhoon. As possible result, catastrophic aftermath such as economic failure, destruction of infrastructures, and loss of lives may be imminent without proper information dissemination and preparedness. Thus, directing all research and technological efforts to disaster preparedness and disaster risk reduction to mitigate the tremendous impacts of natural calamities have been prioritized by the country for years.

According to Statista\footnote{Number of Facebook users in the Philippines: www.statista.com/statistics/490455/number-of-philippines-facebook-users/}, the Philippines has 44 million active Facebook users in 2019 and is predicted to reach approximately 50 million by 2023. Thus, taking in consideration the disaster-prone situation of the country, we note the importance of establishing context-specific knowledge bases where it can be integrated to commonly used digital platforms such as Facebook Messenger to answer context-specific questions on disasters for possible information dissemination and awareness. To make this possible, we present our initial efforts in building a simple knowledge base composed of word concepts joined by semantic word relationships extracted from word embeddings trained from a large online news corpus on disasters. 

The overview of the study is as follows: First, we compared and contrasted previous works done in field of ontology learning and mining information from word embeddings for building knowledge bases. Next, we discussed the method of extraction of concepts and entities from our news article dataset as well as the semantic labels used for building concept relationships. In the Results section, we detail the outcome of using word embeddings trained from our collected dataset for extracting semantically similar concepts using an initial concept seedwords list. In addition, we also discussed the integrity of the disaster ontology from expert validation. We wrap up the study by extensively discussing the semantic changes exhibited by concepts present in the knowledge base over time.

%%
%%
%% --- 2 Previous Work --- %%
%%
%%

\section{Previous Work}
In this section we highlight significant works in using knowledge bases (KBs) as the main source of intelligence for virtual agents as well as current trends using word embeddings as one way of extracting information and semantic word relationships.

\subsection{Building Knowledge-Based Intelligent Agents}
Over the years, the use of powerful knowledge bases are seen on a wide variety of intelligent agents such as chat bots, storytelling agents, and recommender systems. The work of \cite{Ong2018building} focused on building a commonsense knowledge base for a storytelling agent for children using assertions extracted from ConceptNet \cite{Speer2017}. ConceptNet is a large semantic network of word relationships that can be adapted by intelligent agents for commonsense reasoning and identifying object relationships. The study developed a knowledge base by filtering out concepts and concept relationships that are not used in the context of children storytelling from the ConceptNet. 

Similarly, \newcite{Han2015} developed a natural language dialog agent that utilizes a knowledge base to generate diverse but meaningful responses to the user. The system extracts related information from knowledge base, which was adapted from FreeBase \cite{Bollacker2008}, via an information extraction module. The use of a large, external knowledge base allows the dialog system to expand on the information from the users response for a more detailed  and interactive reply.

On the other hand, the work of \newcite{Wang2010} focused on the development of a system composed of three ontology-based sub-agents for personal knowledge, fuzzy inference, and semantic generation for evaluating a person's health through his/her diet. The system makes use of an ontology with an embedded knowledge base considering the persons health statistics such as BMI, Caloric Difference, Health Diet Status combined with rules laid down by domain experts. Results shows that the proposed system exhibits an intelligent behavior in helping the dietary patterns of users based on their information from the constructed ontology.

\subsection{Knowledge from Word Embeddings}
The advent of word embeddings as one of the modern approaches in extracting semantic relationships of words has fueled research works to use its potential to build more powerful knowledge bases. \newcite{Sarkar2018} used a supervised approach, similar to text classification, for predicting the taxonomic relationship via similarity of two concepts using a Word2Vec embedding \cite{Mikolov2013Distributed}. Results showed that combining the word embedding with an SVM classifier outperformed baseline approaches for taxonomic relationship extraction such as using the Jaccard similarity formula and naive string matching.

The study of \newcite{Luu2016} went as far as building a custom neural network architecture with dynamic weighting to significantly increase the performance of statistical and linguistic approaches in extraction word relationships from word embeddings. The neural network considers not only the word relationship such as hypernym and hyponym but also the contextual information between the terms. The proposed approach exhibits generalizability for unseen word pairs and has obtained 9\% to 13\% additional accuracy score using a general and domain-specific datasets.

Likewise, the work of \newcite{Pocostales2016} submitted to the SemEval-2016 Task for Taxonomy Extraction Evaluation \cite{Bordea2016} focused on using GloVe word embedding model \cite{Pennington2014} with an offset feature to extract hypernym candidates from a sample word list. Results showed that a vector offset cannot completely capture the hypernym-hyponym relationship of words due to complexity. 

%
% TABLE - TOP 5 %
% 
\begin{table*}[h]
\begin{center}
\begin{tabular}{ccc}

\begin{tabular}{|c|c|}
\hline \bf 2017 & \bf Tag  \\
\hline \textit{family} & noun \\
\hline \textit{fire} & noun \\
\hline \textit{flood} & noun \\
\hline \textit{update} & verb \\
\hline \textit{tricycle} & noun \\
\hline
\end{tabular}
&
\begin{tabular}{|c|c|}
\hline \bf 2018 & \bf Tag  \\
\hline \textit{act} & verb \\
\hline \textit{announced} & verb \\
\hline \textit{ashfall} & noun \\
\hline \textit{police} & noun \\
\hline \textit{lava} & noun \\
\hline
\end{tabular}
&
\begin{tabular}{|c|c|}
\hline \bf 2019 & \bf Tag  \\
\hline \textit{bulletin} & noun \\
\hline \textit{quake} & noun \\
\hline \textit{typhoon} & noun \\
\hline \textit{weakened} & adjective \\
\hline \textit{affected} & verb \\
\hline
\end{tabular}

\end{tabular}
\end{center}
\caption{ Top 5 seed words per year. }
\label{tab:top5seedwords}
\end{table*}

\section{Data}
For this study, we scraped over 4,500 Filipino disaster-related news articles from years 2017 to 2019 (1,500 articles per year) from Philippine news websites using Octoparse Webscraping Tool as our primary dataset. The corpus covers a wide range of natural disasters that transpired in the Philippines such as  typhoons, earthquakes, landslides and also includes statistics from damages and casualty reports. This large collection of news articles will be the groundwork of the knowledge base as it contains disaster-related context words as concepts. We partitioned the dataset into three by year (2017 to 2019) for the word embedding model generation. The purpose of partitioning the dataset will allow us to analyze the temporal changes of sematic relationships of concepts. More on this is discussed in the succeeding sections. To note, all concepts presented in this document are translated to English for the international audience.

\section{Building the Knowledge Base}

\subsection{Extracting Concept Seedwords}
Building a knowledge base starts with establishing concepts or words that refer to real world entities such as \textbf{nouns}, \textbf{adjectives}, and \textbf{verbs} that represent everyday objects such as \textit{apple, spoon, cake}, people like \textit{mother, police, mayor}, description of objects such as \textit{beautiful, red, big}, and action words that signify an activity such as \textit{walks, eating} and \textit{jumped} \cite{Ong2010commonsense}.  

To identify the grammatical category of each concept to know whether it is a noun, an adjective, or a verb to aid the semantic relationship labelling, we used a Filipino parts-of-speech (POS) tagger\footnote{github.com/matthewgo/FilipinoStanfordPOSTagger} developed by \newcite{Go2017} which is currently integrated in the Stanford CoreNLP package \cite{Manning2014}. We extracted the top 50 high-frequency concepts per year from the collected news dataset, having a total of 150 initial seed words. Table~\ref{tab:top5seedwords} shows the top 5 seed words per year from the initial word list. Both common words such as \textit{family}, \textit{tricycle}, \textit{police} and \textit{update} as well as disaster-related words such as \textit{flood}, \textit{ashfall}, \textit{typhoon}, and \textit{quake} to name a few are present. These concepts are then paired with other concepts to form a meaningful representation of knowledge called a \textbf{binary assertion} described in the next section.

%
% TABLE - RELATION %
% 

\begin{table*}[h]
\begin{center}
\begin{tabular}{|l|l|l|}
\hline \bf Relation & \bf Tag & \bf Rule \\
\hline
Synonym & SYN & If {\tt concept1} has the same meaning with {\tt concept2} \\
Antonym & ANT & If {\tt concept1} has the opposite meaning with {\tt concept2} \\
Hypernym & HYP & If {\tt concept1} has a broader meaning compared to {\tt concept2} \\
Performs & DO & If {\tt concept1} is the actor/does of {\tt concept2} \\
Membership & PartOf & If {\tt concept1} is a member of {\tt concept2} \\
Adjective & IS & If {\tt concept1} describes {\tt concept2} \\
Cause & CAUSE & If {\tt concept1} is the cause of event {\tt concept2} \\
Effect & dueTo & If {\tt concept1} the resulting effect of event {\tt concept2} \\
Random & RAND & If {\tt concept1} has no direct relationship {\tt concept2} \\
\hline
\end{tabular}
\end{center}
\caption{ Semantic relations with its corresponding rules and examples. }
\label{tab:relationsTable}
\end{table*}

\subsection{Semantic Relationship Labelling}
Once the set of context-specific words are obtained from the dataset, in the case of this study, disaster-related concepts, the next process to establish the correct semantic relationship of words. These semantic relationships can be structured in the form of a binary assertion as previously stated. \newcite{Ong2018building} stated that the binary assertions of concepts are needed by virtual agents such as a storytelling agent or a chatbot to be able to generate responses from a commonsense knowledge base. Semantic relationships can also be used for other tasks such query expansion of words as well as information retrieval \cite{Attia2016cogalex}.  For this study, we adapt the binary assertion format used by ConceptNet \cite{Speer2017} shown below:

{\tt \small \textbf{[concept1 semantic-rel concept2]}}

where {\tt \textbf{semantic-rel}} stands for the specified semantic relationship of the two concepts ( {\tt \textbf{concept1}} and  {\tt \textbf{concept2}}) that contains meaning. We adapted the six semantic relation labels from the CogALex-2016 Shared Task on Corpus-Based Identification of Semantic Relations \cite{Santus2016} which are \textbf{Synonym (SYN)} \textbf{Antonym (ANT)}, \textbf{Hypernym (HYP)}, 
\textbf{Membership (PartOf)}, and \textbf{Random (RAND)} as shown in Table~\ref{tab:relationsTable}. In addition, we also added a few semantic labels of our own with respect to the concepts that we will be working on this study which are in the field of disasters. Thus, we added \textbf{Performs (DO)} to signify action, \textbf{Adjective (IS)} to signify description, and \textbf{Cause (CAUSE)} and \textbf{Effect (dueTo)} to signify consequences of events in a disaster setting.

\subsection{Concept Expansion using Word Embeddings}
Word embeddings are representations of an entire document vocabulary in a mathematical vector space. Each word is represented with a set of real-valued numbers given a specific set of dimensions. Word embedding architectures such as \textbf{Word2Vec} \cite{Mikolov2013efficient,Mikolov2013Distributed} and \textbf{Global Vectors} or \textbf{GloVe} \cite{Pennington2014} capture various relationships of words in a corpus such as the \textbf{semantic similarity}, \textbf{syntactic similarity}, and \textbf{word co-occurrence} to name a few. Thus, if two words are commonly used together in the same context, such as in disaster-related articles, we can expect them to be close together when represented in the vector space. For example, finding the most similar terms using the query word \textit{warning} from a word embedding model trained on a disaster-related dataset will output words such as \textit{typhoon}, \textit{flooding}, and \textit{signal} since the word \textit{warning} is commonly used in texts to notify people of possible natural disasters. 

For this study, we generated a word embedding model for each year of partitioned news article dataset from 2017 to 2019. A total of three models were generated using the Word2Vec architecture \cite{Mikolov2013Distributed}. In addition, we used the initial seedwords list which contains 150 concepts (50 for each year) as query words to extract semantically similar terms which occur in the context of disaster.

%
% TABLE - ONTOLOGY EXPANSION %
% 

\begin{table*}[h]
\begin{center}
\begin{tabular}{|l|l|l|}
\hline \bf Seedword & \bf Semantic Label & \bf Assertions \\
\hline
\multirow{3}{*}{police} & DO & [police DO risk]\\
 & IS & [police IS armed]\\
 & DO & [police DO commitment]\\
 \hline
\multirow{3}{*}{tremor} & SYN & [tremor SYN aftershock]\\
 & dueTo & [tremor dueTo quake]\\
 & dueTo & [tremor dueTo earthquake]\\
\hline
\multirow{3}{*}{rescue} & partOf & [rescue partOf operations]\\
 & HYP & [rescue HYP retrieval]\\
 & HYP & [rescue HYP aid]\\
 \hline
 \multirow{3}{*}{experts} & IS & [experts IS supporting]\\
 & DO & [experts DO recommend]\\
 & DO & [experts DO impose]\\
\hline
\end{tabular}
\end{center}
\caption{ Expansion of knowledge base using word embeddings. }
\label{tab:ontologyExpansion}
\end{table*}

\section{Results}
Table~\ref{tab:ontologyExpansion} shows the expanded ontology by querying four sample seedwords from the word embedding model. We only filtered the top three semantically similar resulting words from each query word that falls under a manually-annotated and qualified semantic label discussed in Section 4.2. From this, we arrived at an expanded ontology composed of 450 assertions for our disaster-related knowledge base. 

% FIGURE - Sight Words Acquisition Process

\begin{figure}[b]
    \centering
    \includegraphics[width=0.2\textwidth]{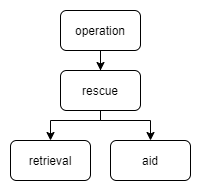}
    \caption{Hierarchy of hypernyms formed from the expanded ontology.}
    \label{fig:rescue}
\end{figure}

From the resulting expanded ontology, it already provides us the \textit{knowledge} that an intelligent agent can piece together or form when asked about something in the context of disasters. Take in, for example, the word \textit{tremor}. Obtaining the top three assertions with semantic labels and expanded using word embeddings informs us that the word \textit{tremor} is synonymous and interchangeable with the word \textit{aftershock} which scientifically means the involuntary movement of the surface due to breaking of underground rocks. Likewise, the words \textit{quake} and \textit{earthquake} are annotated with the semantic label \textit{dueTo} which denotes effect since earthquakes is root cause of tremors according to scientific definition \cite{Yose2013}. 

Another observation from the expanded ontology is the knowledge of understanding general actions to more specific ones. In the example, the action word \textit{rescue} is a \textbf{hypernym}, which means it is a general word that can be possibly specified further, in this case, it is the hypernym of the word \textit{aid}. Consequently, the word \textit{operation} is a hypernym of \textit{rescue}. Combining the words altogether, we can interpret the three assertions as a string of successive actions as shown in Figure~\ref{fig:rescue} where in an \textit{operation} involves a \textit{rescue} and the meaning of \textit{rescue} may vary such that it can be some form of (a) \textit{retrieval} of missing people or (b) \textit{aid} for the wounded or stranded people.

Lastly, the word embedding model was able to produce words that imply \textit{responsibilities} by concepts in the form of human entities. In the two examples of human entities shown in Table~\ref{tab:ontologyExpansion}, \textit{experts} and \textit{police}, most of the semantic labels are \textbf{DO} which denotes \textbf{action} and \textbf{IS} which denotes \textbf{description}. For \textit{police}, common known actions done in the context of disasters are \textit{risk} and \textit{commitment} while being \textit{armed}. For \textit{experts}, the responsibilities are in the line of \textit{supporting} a claim as well as \textit{recommending} and \textit{imposing} future actions based on scientific knowledge. 

We observe the significant potential of using word embeddings for expanding knowledge bases by capturing various levels of information such as (a) understanding interchangeable context-specific words, in the case of this study on disaster-related words, and (b) understanding roles played by human entities described in this section. When used by an intelligent agent, it will have an idea of what response can be produced when queried with questions such as \textit{"What does a police officer do?"} or \textit{"What happens after an earthquake?"}.

\subsection{Expert Validation}
To properly gauge the effectivity of using word embeddings for the expansion of knowledge bases, we performed a validation process by inviting three experts in the fields of linguistics, disaster response, and meteorology to evaluate the assertions of the knowledge base. Each assertion is evaluated using a \textbf{two-scale metric}: \textbf{Agree} if the expert deems that the assertion observes a correct relationship and semantic labelling ([\textit{doctor} SYN \textit{medical person}]) or \textbf{Disagree} if the assertion is not properly labelled to form a correct relationship ([\textit{flood} SYN \textit{typhoon}]).

%
% TABLE - EXPERT VALIDATION %
% 
\begin{table}[h]
\begin{center}
\begin{tabular}{|c|c|}
\hline \bf Model & \bf Aggreability Rate \\
\hline
 Word2Vec 2019 & 0.64 \\
 \hline
 Word2Vec 2018 & 0.52 \\
 \hline
 Word2Vec 2017 & 0.49 \\
\hline
\end{tabular}
\end{center}
\caption{ Expert validation for the knowledge base. }
\label{tab:expertValidation}
\end{table}

Table~\ref{tab:expertValidation} shows the averaged agreeability rate of the expert validation process. Results show that the most accurate model with the highest rating is the Word2Vec model trained on disaster-related news articles collected in the year of 2019. This is followed by word embedding models using the 2018 and 2017 dataset. We attribute the semi-low agreeability scores due to the variation of word usage of the experts in their corresponding fields of study. For example, the assertion [\textit{tremor} SYN \textit{earthquake}] was evaluated by the linguist and disaster experts as \textbf{Agree} while the meteorologist contested. The expert on meteorology swears by the scientific definition of which tremors are very different with earthquakes in a way that tremors are \textit{caused} by earthquakes and not an interchangeable term as perceived by the other two non-technical experts. This pattern is also observed with other assertions using the synonym labelling such as [\textit{storm} SYN \textit{typhoon}] and [\textit{lava} SYN \textit{magma}].

%
% TABLE - ROLES PLAYED BY HUMAN ENTITIES %
% 

\begin{table*}[h]
\begin{center}
\begin{tabular}{|l|l|l|l|}
\hline \bf Seed & \bf 2017 & \bf 2018 & \bf 2019 \\
\hline

\multirow{3}{*}{governor}  & [governor DO assure] & [governor IS political] & [governor DO declare] \\
& [governor DO giving] & [governor DO oversees] & [governor DO resolution] \\
& [governor DO explain] & [governor IS mandatory] & [governor DO develop] \\
  
\hline 

\multirow{3}{*}{experts} & [experts SYN representative] & [experts DO work] & [experts IS supporting] \\
& [experts DO checking] & [experts IS frantic] & [experts DO recommend] \\
& [experts DO verify] & [experts DO announce] & [experts DO impose] \\

\hline

\multirow{3}{*}{police} & [police DO risk] & [police DO risk] & [police DO risk]\\
 & [police DO control] & [police IS armed] & [police DO study]\\
  & [police DO damage] & [police DO commitment] & [police DO alerts]\\
  
\hline
\end{tabular}
\end{center}
\caption{ Roles played by human entities. }
\label{tab:humanEntitiesRoles}
\end{table*}

%
% TABLE - MEMBERSHIPS OF HUMAN ENTITIES %
% 

\begin{table*}[h]
\begin{center}
\begin{tabular}{|l|l|l|l|}
\hline \bf Seed & \bf 2017 & \bf 2018 & \bf 2019 \\
\hline

\multirow{3}{*}{mayor}  & [mayor partOf hall] & [mayor partOf administration] & [mayor RAND workers] \\
& [mayor partOf DWUP] & [mayor DO communication] & [mayor partOf government ] \\
& [mayor partOf sitio] & [mayor partOf DENR] & [mayor RAND employers] \\
\hline 

\multirow{3}{*}{student} & [student partOf PUP] & [student partOf elementary] & [student partOf organization]\\
& [student IS victim] & [student partOf school] & [student partOf University]\\
& [student IS resident] & [student IS minor] & [student partOf UP]\\
\hline 

\multirow{3}{*}{teacher} & [teacher DO education] & [teacher DO research] & [teacher partOf house]\\
& [teacher partOf DepEd] & [teacher DO education] & [teacher SYN employee]\\
& [teacher partOf school] & [teacher partOf school] & [teacher partOf school]\\
\hline 
  
\end{tabular}
\end{center}
\caption{ Memberships of human entities. }
\label{tab:humanEntitiesMembership}
\end{table*}

%
% TABLE - ASSOCIATION OF DISASTER RELATED WORDS %
% 

\begin{table*}[h]
\begin{center}
\begin{tabular}{|l|l|l|l|}
\hline \bf Seed & \bf 2017 & \bf 2018 & \bf 2019 \\
\hline

\multirow{3}{*}{earthquake}  & [earthquake IS magnitude] & [earthquake IS magnitude] & [earthquake IS magnitude] \\
 & [earthquake SYN quake] & [earthquake SYN quake] & [earthquake SYN quake] \\
  & [earthquake RAND drill] & [earthquake IS intensity] & [earthquake RAND signal] \\
\hline 

\multirow{3}{*}{eruption}  & [eruption IS amplifying] & [eruption IS happening] & [eruption IS hazardous] \\
& [eruption IS fast] & [eruption IS explosive] & [eruption IS magmatic] \\
& [eruption IS confirmed] & [eruption CAUSE destruction] & [eruption IS happening] \\
\hline 

\multirow{3}{*}{landslide}  & [landslide dueTo flood] & [landslide dueTo rain] & [landslide RAND mountain] \\
& [landslide HYP mudslide] & [landslide IS torrential] & [landslide dueTo rains] \\
& [landslide RAND area] & [landslide HYP mudslide] & [landslide IS widespread] \\
\hline

\multirow{3}{*}{typhoon}  & [typhoon IS expected] & [typhoon SYN storm] & [typhoon dueTo Amihan] \\
& [typhoon partOf calamity] & [typhoon IS super] & [typhoon SYN hurricane] \\
& [typhoon SYN onslaught] & [typhoon IS powerful] & [typhoon SYN cyclone] \\
\hline 
\end{tabular}
\end{center}
\caption{ Disaster-related terms and its associated words. }
\label{tab:wordAssociationDisaterWords}
\end{table*}

\section{Discussion}
In this section, we perform an even more in-depth analysis by considering the changes in semantic meaning or semantic information of the assertions of the knowledge base over time. We break the discussion into three categories we observed from the analysis.

\subsection{Roles Played by Human Entities}
We observe the changing roles played by three essential human entities, \textit{governor}, \textit{experts}, and \textit{police} over time in the setting of a natural disaster as written in news articles. These entities are expected to be on full alert and their responsibilities are crucial towards mitigating the consequences of disasters. 

The entity \textit{governor}, or the highest commanding individual of Philippine province, performs various roles over time. As seen in Table~\ref{tab:humanEntitiesRoles}, the entity is mostly connected by the semantic label \textbf{DO} to denote action with concepts such as \textit{assure}, \textit{explain}, \textit{oversee}, \textit{develop}, and \textit{declare}. There are also a few description words connected by the label \textbf{IS} such as \textit{political} and \textit{mandatory} which is obvious since holding a gubernatorial position is indeed \textit{political} and resolutions filed in a governor's office is in its essence \textit{mandatory}. 

Similarly, the entity \textit{experts} is also expected to have various changing roles. For 2017 as seen in the Table, common action words associated are \textit{checking} and \textit{verifying} which provides one of the most important roles of experts in the field of disasters: validating the integrity of information being publicized. For 2018, experts play more of an information dissemination role with the concept \textit{announce} as the connecting action word. In 2019, experts assumed a more stricter role as observed with the connected concepts such as \textit{recommend} and \textit{impose}.

For the entity \textit{police}, there is consistent associated descriptor across the years regarding their responsibility: \textit{risk}. This provides us a concrete idea of a consequence when assuming the role of a police. In 2017 and 2018, strong action words such as \textit{control}, \textit{damage}, and \textit{armed} are tied with a policeman's job. In 2019, however, the entity \textit{police} assumed a more passive role as more of an informant with the connected action words being \textit{study} and \textit{alerts}.

\subsection{Memberships of Human Entities}
We also observed changes in memberships of human entities in context of disasters over time as seen in Table~\ref{tab:humanEntitiesMembership}. Memberships are denoted by the semantic labels \textbf{HYP} for \textbf{hypernyms} or general words and \textbf{partOf} for \textbf{hyponyms} more specific words. We observe memberships played of three entities: \textit{mayor}, \textit{student}, \textit{teacher}.

For the entity \textit{mayor}, memberships are more specific in 2017 and 2018. The entity is expected to partner with small groups in a municipality such as a \textit{sitio} or a barangay cite as well as with large government agencies such as Division on the Welfare of the Urban Poor (DWUP) and Department of Environment and Natural Resources (DENR) in times of disasters. There are also general memberships such as \textit{administration} and \textit{government} to which the entity is an obvious member.

In the case of the entity \textit{student}, there is a mix of specific and generalized memberships. Certain specific universities such as \textit{UP} or the University of the Philippines and \textit{PUP} or Polytechnic University of the Philippines were classified as institutions where a student may belong. General and obvious concepts such as \textit{elementary}, \textit{school}, \textit{organization} and \textit{University} contribute to the commonsense information of the knowledge base.

For the entity \textit{teacher}, general concepts of memberships are more prominent compared to the other two entities. The concept \textit{school} is consistent for all years. The term \textit{DepEd} which means Department of Education, the government agency responsible in shaping the educational landscape of the Philippines, is connected to the entity. DepEd oversees elementary and intermediate level schools which may mean that the term \textit{teacher} is commonly tied with educators from these levels. Interestingly, the concept \textit{house} is also tied with the entity \textit{teacher}. Although it may already be obvious that a teacher assumes a different role inside the house maybe as a mother or a breadwinner.

\subsection{Common Word Association of Disaster-Related Terms}
For the last category, we observe change in association of disaster-related words over time as seen in Table~\ref{tab:wordAssociationDisaterWords}. We highlight the importance of this analysis to understand how a knowledge base using information extracted from word embeddings interchange co-occurring and similar words in the context of disasters. For this category, we analyze the four frequently occurring natural disasters in the Philippines which are \textbf{earthquakes}, \textbf{eruption}, \textbf{landslide}, and \textbf{typhoon}.

The first disaster-related word is \textit{earthquake}. From the formed assertions for all years, \textit{earthquakes} are synonymous with the term \textit{quakes} which denote a shortened version of the word. In addition, the term \textit{magnitude} is also a consistent term connected to the seedword \textit{earthquake} which tells us that earthquakes have their own corresponding \textit{magnitudes} quantified by some number. 

In the case of \textit{eruption}, most assertions formed are used with the semantic label \textbf{IS} to denote \textbf{description} or \textbf{property}. Across the years, most of the descriptive words associated with the term are negative such as \textit{amplifying}, \textit{hazardous}, \textit{destruction}, \textit{explosive}, and \textit{magmatic}. These word may denote a sense of urgency compared to the other natural disasters when it happens. The term \textit{happening} has occurred both in 2018 and 2019 due to two of the most active volcanoes in the Philippines, Mount Mayon and Mount Taal, had shown activity\footnote{Volcano Bulletin: phivolcs.dost.gov.ph/index.php/volcano-hazard/volcano-bulletins3} by erupting successively and spewing ash. The eruption caused mass postponement of public activities and over 48,000 evacuated locals.

For the concept \textit{landslide}, the common associated word is \textit{mudslide}. Although by scientific definition, mudslides are as specific type landslides (also called \textit{debris flow}). Thus, the \textit{landslide} term conforms the the semantic label \textbf{HYP} for hypernyms. The cause of landslides can be tied to \textit{floods}, which in turn, caused by \textit{rains} as joined by the semantic label \textbf{dueTo} denoting a \textbf{consequence} or an \textbf{effect}. In addition, landslides are also often described using the words \textit{widespread} and \textit{torrential} which gives us a quantifiable idea of the magnitude of landslides that occur in the Philippines.

Lastly, we have the word \textit{typhoon}. This disaster-related concept assumes many interchangeable and synonymous terms such as \textit{storm}, \textit{hurricane}, \textit{cyclone}, and \textit{onslaught}. We note the frequent association and interchangeability of these words due to geographic locations \cite{Khadka2018} but may mean the same thing---they are all \textbf{tropical storms}. Likewise, description words tied to the concept \textit{typhoon} are \textit{expected}, \textit{super}, and \textit{powerful} which also provides us an idea of the magnitude of typhoons occurring in the country similar to \textit{landslides}.

\section{Conclusion}
As a disaster-prone country, the Philippines should increase its efforts in mitigating the effects of natural calamities with the help of technology. One way to do this is to consider the potential of intelligent agents such as chatbots as tools for disaster preparedness and information dissemination. In this study, we established a simple context-specific knowledge base by doing three important processes: (a) extracting disaster-related concepts from a collected news article dataset, (b) building a network of binary assertions from a curated list of semantic labels, and (c) expanding the ontology by querying an initial seedwords list from word embeddings generated from the original news dataset. Results show that using word embeddings captured various levels of information that may be useful for intelligent agents to produce responses such as information on roles of human entities, generalization and specification of terms, and common word association when asked in the topic of disasters. Future directions of the study include collection of even more dataset that covers not only news articles but also other media for finer-grained assertions. In addition, the study will also benefit from efforts in testing the capability of the knowledge base in practical applications.

\bibliographystyle{acl}
\bibliography{ramirez_references}

\end{document}